\pdfoutput=1
\documentclass[conference]{IEEEtran}
\IEEEoverridecommandlockouts

\usepackage{cite}
\usepackage{amsmath,amssymb,amsfonts}
\usepackage{graphicx}
\usepackage{booktabs}
\usepackage{textcomp}
\usepackage{xcolor}
\usepackage{tikz}
\usetikzlibrary{arrows.meta, positioning, fit}

\newcommand{\ci}[2]{\,{\scriptsize[#1,\,#2]}}

\begin{document}

\title{A Reverse Sign Language Dictionary:\\
Open-Vocabulary Sign Recognition from Continuous Signing\\
via Video Captioning and Description Retrieval}

\author{
\IEEEauthorblockN{Santiago Poveda-Guti\'errez}
\IEEEauthorblockA{\textit{Grad.\ School of IST} \\
The University of Tokyo \\
Tokyo, Japan \\
santiago@nlab.ci.i.u-tokyo.ac.jp}
\and
\IEEEauthorblockN{{Hideki Nakayama}}
\IEEEauthorblockA{{\textit{Grad.\ School of IST}} \\
{The University of Tokyo} \\
{Tokyo, Japan} \\
{nakayama@ci.i.u-tokyo.ac.jp}}
\and
\IEEEauthorblockN{{Mayumi Bono}}
\IEEEauthorblockA{{\textit{Information and Society Research Division}} \\
{National Institute of Informatics} \\
{Tokyo, Japan} \\
{bono@nii.ac.jp}}
}

\maketitle
\begin{abstract}
Isolated Sign Language Recognition (ISLR) is conventionally cast as closed-set
classification over gloss labels, which cannot generalize to signs unseen in
training and ties every deployment to a gloss-annotated lexicon. We instead
recognize signs extracted from \emph{continuous} signing by (1) captioning a
sign-level clip into a free-form procedural description of the articulation
with an open-weight vision--language model, and (2) retrieving the closest
entry from a vocabulary of target descriptions with a multilingual sentence
encoder: a \emph{reverse sign language dictionary} that needs no gloss
supervision and admits an open vocabulary. On 1{,}300 sign-level segments from
a Japanese Sign Language (JSL) dialogue corpus annotated with procedural
descriptions (against a $2\%$ top-10 chance floor over the 503-entry target
vocabulary), fine-tuning the captioner substantially improves seen-class
retrieval: language and vision tower fine-tuning raises top-10
retrieval on seen classes from $4.5\%$ (untrained) to $49\%$, becoming statistically indistinguishable from a standard supervised closed-set classifier (I3D) on two of the three test sets where a closed-set classifier can be evaluated at all. More importantly, unseen-class retrieval also improves significantly over the
untrained pipeline ($11.5\%\!\to\!21.0\%$ top-10, $p=0.0094$), a regime in
which the closed-set classifier cannot participate. A matcher-side
empirical upper-bound analysis shows the sentence encoder already recovers
close to $100\%$ of paraphrased gold descriptions, locating a gap in
captioning quality that we aim to address in future work. To
our knowledge this is the first description-based, open-vocabulary sign
lookup from continuous signing without gloss supervision, and the first for
JSL.
\end{abstract}


\section{Introduction}

Signed languages are complete natural languages, and ISLR underpins
accessibility tools such as sign-language dictionaries, where lookup means
finding the meaning of a sign one just saw. Yet most ISLR research assumes
isolated, deliberately produced clips and a fixed gloss inventory. Naturally occurring signing, on the contrary, is continuous (coarticulated, fast, $\approx$0.5\,s/sign
on average in our corpus), and the lexicon is never fully covered by training.

Recognizing signs within continuous signing exists in the literature as \emph{sign
spotting}~\cite{momeni2020watch,albanie2020bsl1k}, but it is gloss-supervised and
closed-vocabulary. Zero-shot recognition from textual sign descriptions has been
studied for isolated, dictionary-style productions~\cite{bilge2023zsslr}, and
dictionary-retrieval framings of ISLR~\cite{desai2023aslcitizen,jiang2024signclip}
are the closest neighbors of our pipeline. To our knowledge, however, no prior
method performs \textbf{description-based, open-vocabulary sign lookup from
continuous signing without gloss supervision}, and none of the sign-spotting
literature targets JSL. Our research questions are deliberately basic:
\emph{can this concept work in practice?}, and \emph{can it be trained while
still retrieving signs never seen in training?} We contribute (1) a modular
captioning + retrieval pipeline for open-vocabulary ISLR from continuous
signing; (2) an evaluation protocol with zero-shot unseen-class and
view-generalization splits against a closed-set baseline and a matcher
upper bound; and (3) a controlled fine-tuning study across two regimes
(captioner SFT with the vision tower frozen or also adapted), showing that
adapting the vision tower yields the larger improvement and also moves unseen-class retrieval significantly above
the untrained baseline.

\section{Method}

\subsection{Pipeline}
\begin{figure}[t]
\centering
\resizebox{0.98\columnwidth}{!}{%
\begin{tikzpicture}[
  font=\sffamily\footnotesize,
  box/.style={draw, rounded corners=2pt, align=center, minimum height=0.62cm, inner sep=3pt},
  arr/.style={-{Latex[length=2mm]}, thick},
  node distance=6mm
]
  \node[box, fill=blue!6] (clip) {Sign-level\\clip};
  \node[box, fill=orange!12, right=of clip, minimum width=1.7cm] (cap) {Captioner\\(LVLM)};
  \node[box, fill=blue!6, right=of cap, minimum width=1.9cm] (desc) {Generated\\description};
  \node[box, fill=orange!12, right=of desc, minimum width=1.7cm] (match) {Matcher\\(sent.\ encoder)};
  \node[box, fill=blue!6, right=of match, minimum width=1.6cm] (rank) {Top-$k$\\ranked list};

  \draw[arr] (clip) -- (cap);
  \draw[arr] (cap) -- (desc);
  \draw[arr] (desc) -- (match);
  \draw[arr] (match) -- (rank);

  \node[box, dashed, fill=gray!8, below=9mm of desc, minimum width=2.5cm] (vocab)
    {Target vocabulary\\(503 descriptions)};
  \draw[arr] (vocab.north) -- (match.south);

  \node[draw, dotted, thick, color=red!70!black, fit=(cap), inner sep=4pt,
        label={[red!70!black, font=\sffamily\scriptsize]above:{\shortstack{fine-tuned:\\LM LoRA (+ ViT LoRA)}}}] {};
\end{tikzpicture}}
\caption{The pipeline. Both stages are swappable; fine-tuning (\S\ref{sec:training})
updates the captioner's language-model adapters and, in the vision-adapted
regime, a low-rank adapter on its vision tower as well. The matcher and
vocabulary are untouched throughout the results reported here.}
\label{fig:pipeline}
\end{figure}
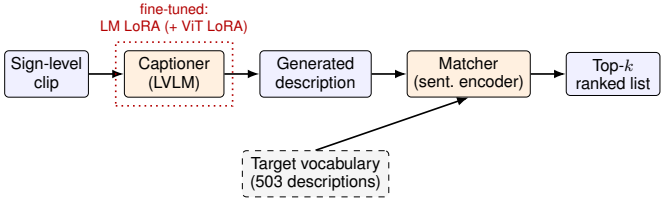

Given a sign-level clip, an open-weight large vision--language model (LVLM;
InternVL3, 8B~\cite{internvl3}) is prompted to produce a procedural description: how the sign
is performed (e.g., hand shape, location, movement). The produced description is embedded
by a multilingual sentence encoder (BGE-M3~\cite{chen2024bgem3}) and scored
by cosine similarity against the precomputed target vocabulary, returning sorted top-$k$
entries (Fig.~\ref{fig:pipeline}). Both stages are swappable
modules behind a common interface; captioning and matching are performed
in Japanese throughout this paper (\S\ref{sec:data}), the language of the
ground-truth descriptions.

\subsection{Data and evaluation protocol}
\label{sec:data}
We use a subset of the JSL dialogue corpus~\cite{bono2014colloquial} whose sign-level segments are annotated with \emph{Relevant Annotation}~\cite{bono2016annotation}:
free-form Japanese descriptions of how each sign is articulated in
conversation, which often differ from its dictionary citation form. It comprises
1{,}300 segments, 5 signers (2 prefectures),
semi-frontal and side views, and 503 unique descriptions. The two active-signer single-view crops of each segment
are used for testing. Each query has exactly one gold description; we
report \textbf{top-$k$ retrieval accuracy} ($k\!=\!1,5,10$) against the full
vocabulary, with Wilson 95\% CIs. Out of the corpus, we define four test sets: \textbf{TS1} (unseen
segments, seen classes); \textbf{TS2} (singleton descriptions, zero training
segments, inaccessible to closed-set classifiers); \textbf{TS3.1/TS3.2} (view
generalization,
semi-frontal$\leftrightarrow$side, sharing the same 100 descriptions/segments
so class difficulty is not a confound). We compare against a supervised
closed-set classifier (I3D, an Inflated 3D ConvNet~\cite{carreira2017i3d},
initialized on Kinetics, a large action-recognition video dataset, and
fine-tuned per test set~\cite{li2020wlasl}; a standard closed-set reference
point) and a matcher upper bound: an
LLM produces three types of paraphrases for each gold description (lexical,
structural, and free-form rewrites, the last closest to how a real caption
could read), which are fed to the matcher directly, isolating
matcher error from captioning error.

\subsection{Fine-tuning: two regimes}
\label{sec:training}
Both regimes fine-tune InternVL3-8B with LoRA~\cite{hu2022iclr-lora} via LLaMA-Factory~\cite{zheng2024llamafactory}. In both, we use one uninterrupted
9-epoch cosine schedule, per-test-set training pools with the same train/test
exclusion rules, and identical prompts/decoding to the off-the-shelf pipeline.
\textbf{Regime A} (frozen vision tower) trains only the language model's
LoRA adapters and the multimodal projector. \textbf{Regime B}
(vision-adapted) additionally adds the vision transformer (ViT)'s linear layers to
the LoRA target set. Each test set's A/B pair is a \emph{matched control}
(regime A and B are otherwise identical runs, so any difference between
them isolates one variable): identical data, schedule, and seed, differing
only in whether the vision tower is adapted.

\section{Results}

\begin{table*}[t]
\centering
\caption{Retrieval accuracy (\%) by training regime, with reference points, all
InternVL3-8B/JA unless noted (Wilson 95\% CIs in brackets). Ceiling is the
matcher upper bound (\S\ref{sec:data}), averaged over the three paraphrase
types. Bold: the best of the four competing methods (off-the-shelf,
+SFT frozen ViT, +SFT adapted ViT, I3D) per row. Chance floor: top-1/5/10
$0.2/1.0/2.0\%$ respectively (503-entry vocabulary) at every $n$ below.}
\label{tab:main}
\scriptsize
\renewcommand{\arraystretch}{0.72}
\setlength{\tabcolsep}{4pt}
\begin{tabular}{@{}llccccc@{}}
\toprule
Test set ($n$) & $k$ & Off-the-shelf & +SFT (frozen ViT) & +SFT (adapted ViT) & I3D & Ceiling \\
\midrule
TS1, seen (200) & 1  & 0.5\ci{0.1}{2.8}   & 18.5\ci{13.7}{24.5} & \textbf{29.0}\ci{23.2}{35.6} & 15.5\ci{11.1}{21.2} & 89.7\ci{87.0}{91.9} \\
 & 5  & 3.5\ci{1.7}{7.1}   & 28.5\ci{22.7}{35.1} & \textbf{40.0}\ci{33.5}{46.9} & 39.0\ci{32.5}{45.9} & 98.7\ci{97.4}{99.3} \\
 & 10 & 4.5\ci{2.4}{8.3}   & 36.0\ci{29.7}{42.9} & 49.0\ci{42.2}{55.9} & \textbf{52.0}\ci{45.1}{58.8} & 100.0\ci{99.4}{100.0} \\
\midrule
TS2, unseen (200) & 1  & \textbf{2.5}\ci{1.1}{5.7} & 0.0\ci{0.0}{1.9}  & 0.0\ci{0.0}{1.9}   & N/A & 92.0\ci{89.6}{93.9} \\
 & 5  & 7.5\ci{4.6}{12.0}  & 10.5\ci{7.0}{15.5}  & \textbf{13.5}\ci{9.5}{18.9}  & N/A & 98.7\ci{97.4}{99.3} \\
 & 10 & 11.5\ci{7.8}{16.7} & 16.5\ci{12.0}{22.3} & \textbf{21.0}\ci{15.9}{27.2} & N/A & 99.7\ci{98.8}{99.9} \\
\midrule
TS3.1, $\to$semifr. (100) & 1  & 0.0\ci{0.0}{3.7}   & 19.0\ci{12.5}{27.8} & \textbf{34.0}\ci{25.5}{43.7} & 12.0\ci{7.0}{19.8}  & 90.7\ci{86.8}{93.5} \\
 & 5  & 6.0\ci{2.8}{12.5}  & 29.0\ci{21.0}{38.5} & \textbf{39.0}\ci{30.0}{48.8} & 27.0\ci{19.3}{36.4} & 97.7\ci{95.3}{98.9} \\
 & 10 & 6.0\ci{2.8}{12.5}  & 36.0\ci{27.3}{45.8} & 45.0\ci{35.6}{54.8} & \textbf{46.0}\ci{36.6}{55.7} & 99.3\ci{97.6}{99.8} \\
\midrule
TS3.2, $\to$side (100) & 1  & 1.0\ci{0.2}{5.5}   & 16.0\ci{10.1}{24.4} & \textbf{27.0}\ci{19.3}{36.4} & 14.0\ci{8.5}{22.1}  & 90.7\ci{86.8}{93.5} \\
 & 5  & 6.0\ci{2.8}{12.5}  & 22.0\ci{15.0}{31.1} & 36.0\ci{27.3}{45.8}          & \textbf{61.0}\ci{51.2}{70.0} & 97.7\ci{95.3}{98.9} \\
 & 10 & 7.0\ci{3.4}{13.7}  & 27.0\ci{19.3}{36.4} & 40.0\ci{30.9}{49.8}          & \textbf{75.0}\ci{65.7}{82.5} & 99.3\ci{97.6}{99.8} \\
\bottomrule
\end{tabular}
\end{table*}

\begin{figure}[t]
\centering
\includegraphics[width=0.98\columnwidth]{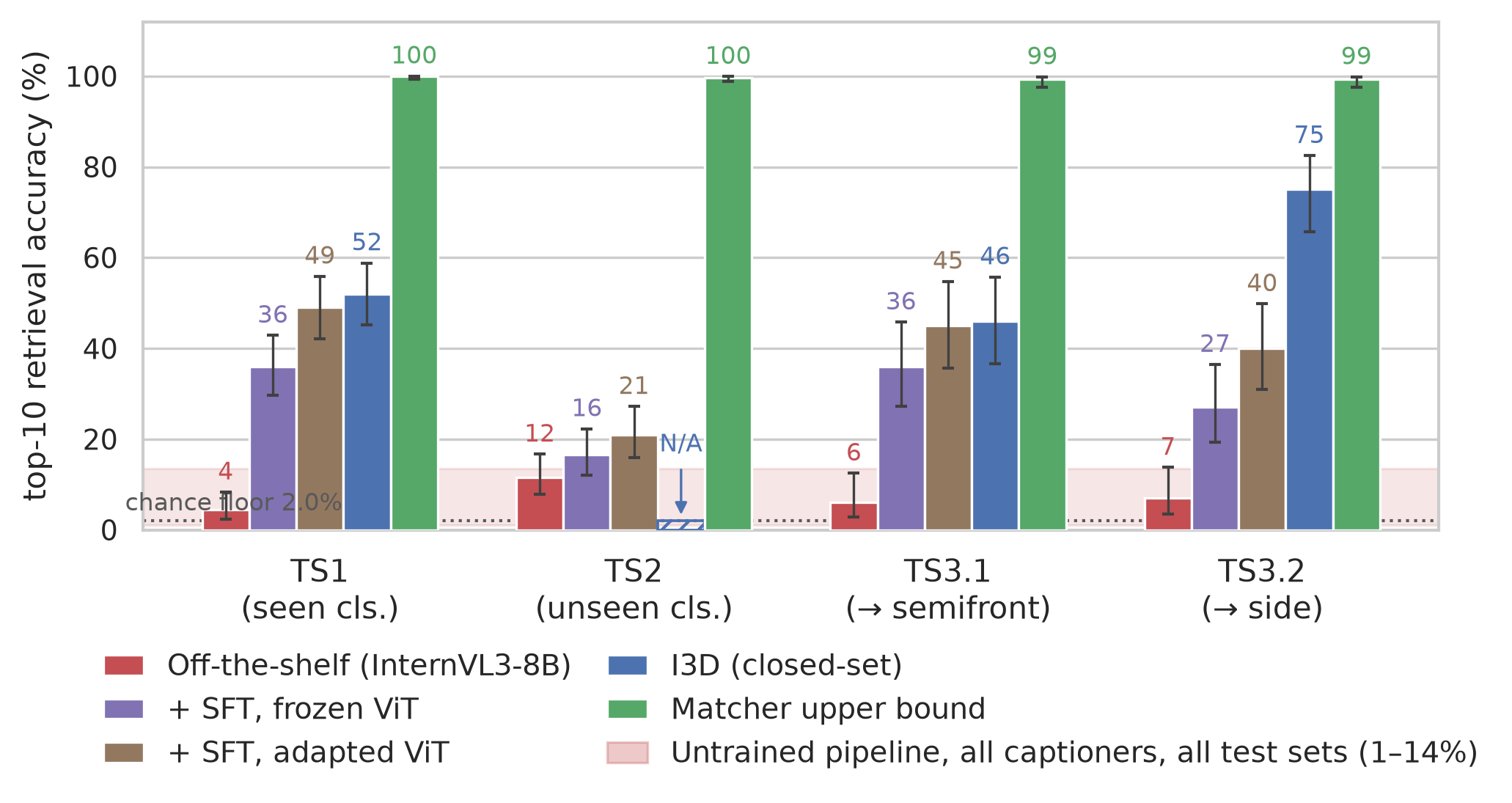}
\caption{Top-10 accuracy comparison, error bars represent Wilson 95\% CIs.
Left to right per test set: our three regimes, I3D (closed-set; N/A on TS2),
and the matcher upper bound (retrieval average over the three paraphrase
types defined in \S\ref{sec:data}). The untrained band is the
global min--max top-10 accuracy pooled across test sets over every open-weight LVLM we screened
off-the-shelf as a captioner (InternVL3-8B, used elsewhere in this paper,
plus several other open-weight vision-language models) in this same
Japanese matching configuration. See
\S\ref{sec:results-main} for the per-test-set comparison this figure
summarizes.}
\label{fig:landscape}
\end{figure}

\subsection{Vision-adapted fine-tuning matches I3D at top-5/top-10, beats it
at top-1, and improves unseen-class retrieval}
\label{sec:results-main}

Table~\ref{tab:main} and Fig.~\ref{fig:landscape} illustrate the results: each
step (off-the-shelf$\rightarrow$frozen-ViT SFT$\rightarrow$vision-adapted SFT) improves
retrieval on \emph{every} test set, seen and unseen alike. Table~\ref{tab:main}
itself reports only point estimates and Wilson CIs; every $p$-value in this
section, starting here, is instead from a separate paired exact McNemar test
on the same items, comparing two of the table's accuracy figures at a time.
On TS1, adapting the vision tower adds $+13.0$pp top-10 over the frozen-ViT
control ($p=8.6\!\times\!10^{-4}$), $+44.5$pp over untrained
($p=2.4\!\times\!10^{-22}$). The rest of this section reports three further
findings against I3D and the untrained pipeline.

\noindent\textbf{Top-5/top-10 retrieval is now comparable to I3D on two of
three eligible test sets.} On \textbf{TS1}, top-5/top-10 are a statistical
draw ($p=0.90$, $p=0.55$). \textbf{TS3.1} goes further: the pipeline leads
or draws every cell (top-5 $p=0.073$, top-10 $p=1.0$). \textbf{TS3.2}
reverses it: I3D keeps a significant lead at top-5/top-10 ($p=0.00025$,
$p=1.2\!\times\!10^{-6}$). We ruled out train/test leakage as the cause, and the training-pool composition is structurally
symmetric between TS3.1 and TS3.2, so it could not explain an asymmetry
between them. We do not yet have an explanation for why this specific test
direction (predicting the side view) favors the closed-set classifier; it
may be a particularity of this I3D instantiation.

\noindent\textbf{Top-1 accuracy is higher than I3D on every eligible test
set.} The pipeline leads I3D at top-1 on TS1 ($29.0$ vs.\ $15.5$,
$p=6.6\!\times\!10^{-5}$), TS3.1 ($34.0$ vs.\ $12.0$,
$p=2.7\!\times\!10^{-5}$), and TS3.2 ($27.0$ vs.\ $14.0$, $p=0.029$) alike,
including on the one test set where I3D wins at top-5/top-10. A closed-set
classifier cannot appear in a TS2 row at all: it has no output unit for a
class with zero training segments, so it is confined to the chance floor by
construction there.

\noindent\textbf{Training improves unseen-class retrieval.} Frozen-ViT SFT
alone leaves TS2 statistically flat relative to untrained. With the vision
tower adapted, TS2 top-10 improves significantly over untrained
($11.5\%\!\to\!21.0\%$, $p=0.0094$). Stated precisely: the gain is
significant \emph{against untrained}; against the frozen-ViT matched
control it is directionally positive but not significant at $n=200$
($+4.5$pp, $p=0.122$), so its attribution to the vision tower specifically
remains directional.

\subsection{Why LoRA-on-ViT instead of full ViT fine-tuning}
\label{sec:mechanism}

We conducted an ablation that rules out ``more visual parameters is simply
better'' \emph{at this data scale}, not in general: a full fine-tune of the
entire vision tower at the same learning rate collapses the model (one
distinct caption across 200 clips at epoch~1). A
$2\times2$ (LoRA-on-ViT vs.\ full ViT $\times$ two learning rates) shows that
capacity helps at matched learning rate ($+9.0$pp, $p=0.025$), but the lower
learning rate a full fine-tune needs costs more than that gain ($-15.0$pp,
$p\!=\!1.0\!\times\!10^{-4}$), since it also slows the language model's own
adapters. LoRA-on-ViT at a moderate, matched learning rate remains the best
recipe at this data scale ($\approx$800 unique training segments), evidence
for a data-limited ceiling, motivating the large-scale masked video
pre-training proposed under ``Longer-horizon'' below (\S\ref{sec:limitations}).

The matcher is not the limiting factor: paraphrasing the gold description
recovers it $90$--$100\%$ of the time (Table~\ref{tab:main}'s Ceiling
column), so the remaining gap sits concretely in what the captioner sees
and says, not in retrieval.

\section{Limitations and Future Work}
\label{sec:limitations}

\noindent\textbf{Limitations.} Each configuration is trained once (a single
random seed), so we cannot yet separate a genuine training effect from
ordinary run-to-run variance. Sample sizes are modest ($n=100$--$200$):
across Table~\ref{tab:main}'s 36 cells the average Wilson CI half-width is
$\pm5.7$pp, which bounds how small an effect this design can resolve in
general, including the TS2 arm-vs-control comparison above. All 5 signers
appear in both training and test. Held-out test segments, TS2's singletons
included, are drawn from longer recording sessions that also contain
training segments (of other classes), so some trained accuracy on any test
set may partly reflect that shared context (signer, clothing, background)
rather than only sign recognition transferring. The corpus remains
small and single-language.

Separately, the fine-tuned captioner's output collapses toward its own
training vocabulary: on TS1, over 90\% of its generated captions are
near-verbatim copies of a training-set target description, versus 0\%
before fine-tuning, and this rate is statistically unchanged whether or not
the vision tower is adapted. We read this as a signature of overfitting at
the current data scale (a few hundred to a few thousand training clips per
test set) rather than a property of vision-tower adaptation specifically,
and interpret it as a further argument for the large-scale pre-training proposed
below.

This also explains an apparent tension in \S\ref{sec:results-main} above:
training raises TS2 top-5/top-10 while top-1 stays at $0.0\%$ for both
regimes (Table~\ref{tab:main}). Since TS2 classes have zero training
segments of their own, a verbatim copy of a training-set description can
never be the exact gold string, so top-1 is structurally $0\%$ once the
captioner is collapsing onto that vocabulary. Top-5/top-10 still improve
because the collapsed captions the model reaches for are semantically close
to the true gold, close enough for the matcher to place the gold in the
shortlist, which is itself evidence that visual understanding is
happening as well.

\noindent\textbf{Planned next.} (1) A cross-lingual test set, testing
whether the open-vocabulary property transfers across languages as well as
across classes. (2) A CLIP-style contrastive video-text baseline, to test
whether a different architecture generalizes as well to unseen sentence
formats and target dictionaries.

\noindent\textbf{Longer-horizon.} Large-scale masked video pre-training of the
vision tower before captioner SFT, directly motivated by
\S\ref{sec:mechanism}; paraphrase-augmented SFT training targets, to address
the output-space collapse; a video-aware reranker and paraphrase-aware matcher.

\section*{Acknowledgments}

We thank Kazuyoshi Yoshii, Tatsuya Kawahara, Junwen Mo, and Duc Minh Vo for
valuable discussions on the pipeline and evaluation protocol, and the Deaf
annotators for their time and expertise.

\textbf{Funding:} This work was supported by JSPS KAKENHI Grant Numbers 22B102 and 22H05014, by the Japan Science and Technology Agency (JST) under the Adopting Sustainable Partnerships for Innovative Research Ecosystem (ASPIRE) program, Grant Number JPMJAP25B3, and by the Mitsubishi Zaidan Foundation, project ID 202420002, ``Understanding Minority Languages and Communication Using Artificial Intelligence Technology.''

\bibliographystyle{IEEEtran}
\bibliography{references}

\typeout{get arXiv to do 4 passes: Label(s) may have changed. Rerun}
\end{document}